\documentclass[11pt,onecolumn]{article}  

\usepackage{graphicx}
\usepackage{amssymb}
\usepackage{epstopdf}
\usepackage{graphics}
\usepackage{amsmath, amssymb}
\numberwithin{equation}{section}
\usepackage{graphicx}   
\usepackage{verbatim}   
\usepackage{color}      
\usepackage{subfigure}  
\usepackage{hyperref}   
\usepackage{xspace}
\usepackage{float}
\usepackage{amsthm}
\usepackage{fancyhdr,lastpage}
\usepackage{lastpage}
\usepackage{wrapfig}
\usepackage[font={footnotesize}]{caption}

\usepackage{sidecap}
\definecolor{gray}{RGB}{190,190,190}

\DeclareMathOperator*{\argmax}{arg\,max}

\newcommand{\bff}{\mathbf{f}}

\newcommand{\bfz}{\mathbf{z}}

\newcommand{\bs}[1]{\boldsymbol #1}

\newcommand{\boeta}{\bs\eta}
\newcommand{\bfr}{\bff^R}

\newcommand{\probot}{p(\bff^{R}\mid\bfz^{R}_{1:t})}

\newcommand{\pigp}{p_{}(\bff^{R},\bff^{1},\ldots,\bff^{n}\mid\bfz_{1:t})}
\newcommand{\pigpshort}{p(\bff^{R},\bff\mid\mathbf z_{1:t})}
\newcommand{\na}{p(\boeta_0,\bff^{(R)},\bff\mid\bfz_{1:t})}
\newcommand{\pzero}{p(\boeta_0 \mid  \{ G_i\}_{i=1}^m)}

\title{\LARGE \bf
Integrating High Level and Low Level Planning}
  
\author{Pete Trautman
}

\date{}

\begin{document}

\maketitle
\thispagestyle{empty}
\pagestyle{empty}

\begin{abstract}
\noindent We present a possible method for integrating high level and low level planning.  To do so, we introduce the global plan random \emph{trajectory} $\boldsymbol{\eta}_0 \colon [1,T] \to \mathbb R^2$, measured by goals $G_i$ and governed by the distribution $p(\boldsymbol{\eta}_0 \mid  \{ G_i\}_{i=1}^m)$.  This distribution is combined with the low level robot-crowd planner $p(\mathbf{f}^{R},\mathbf{f}^{1},\ldots,\mathbf{f}^{n}\mid\mathbf{z}_{1:t})$ (from~\cite{trautmanicra2013, trautmaniros}) in the distribution $p(\boldsymbol{\eta}_0,\mathbf{f}^{(R)},\mathbf{f}\mid\mathbf{z}_{1:t})$.   We explore this \emph{integrated planning} formulation in three case studies, and in the process find that this formulation 1) generalizes the ROS navigation stack in a practically useful way 2) arbitrates between high and low level decision making in a statistically sound manner when unanticipated local disturbances arise and 3) enables the integration of an onboard operator providing real time input at either the global (e.g., waypoint designation) or local (e.g., joystick) level.  Importantly, the integrated planning formulation $p(\boldsymbol{\eta}_0,\mathbf{f}^{(R)},\mathbf{f}\mid\mathbf{z}_{1:t})$ highlights failure modes of the ROS navigation stack (and thus for standard hierarchical planning architectures); these failure modes are resolved by using $p(\boldsymbol{\eta}_0,\mathbf{f}^{(R)},\mathbf{f}\mid\mathbf{z}_{1:t})$. Finally, we conclude with a discussion of how results from formal methods can guide our factorization of $p(\boldsymbol{\eta}_0,\mathbf{f}^{(R)},\mathbf{f}\mid\mathbf{z}_{1:t})$.
\end{abstract}

\section{Case Studies}
To explain our intended approach to integrating high and low level planning, we introduce the high level motion plan random \emph{trajectory} variable $\boeta_0$ that is governed by the distribution $\pzero$ and conditioned on symbolic data $\{G_i \}_{i=1}^m$.  We treat the high level plan as a random variable because of the following: the high level planner must be able to accommodate local disturbances returned by the low level motion planner.  In turn, high level motion plans must be able to adjust to online goal changes; these high level changes must then trickle down to low level behavior.  Conceptually then, the high level plan and the low level plan are \emph{coupled} variables; if either is restricted to a single hypothesis (as is typical in conventional approaches to hierarchical planning), then the high and low level plans are unable to influence each other.   

\begin{wrapfigure}{r}{0.33\textwidth}
\centering \vspace{-1em}
    \includegraphics[width=0.3\textwidth]{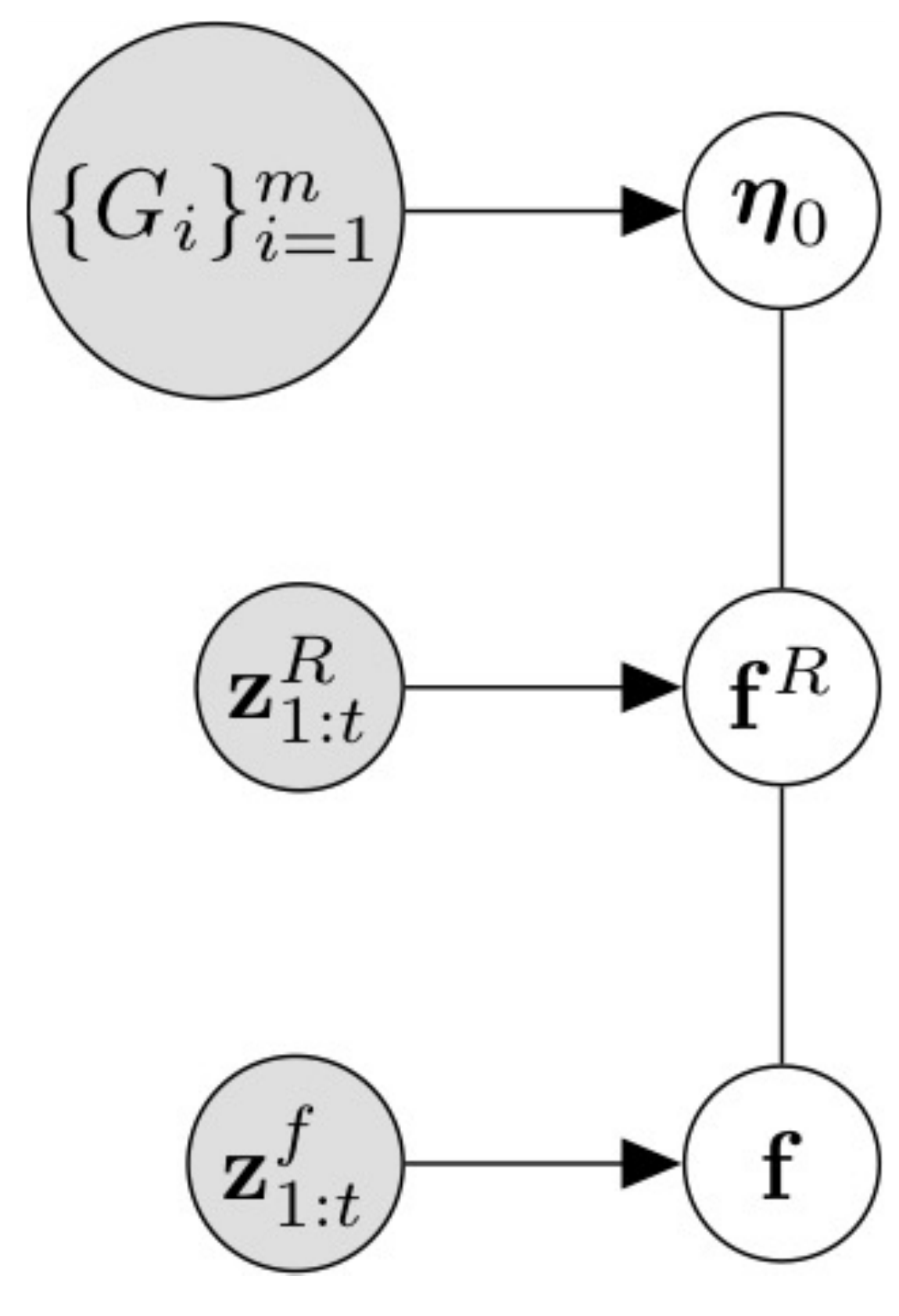}
    \vspace{-1em}
    \caption{Graphical model depicting the relationship between high level plans $\boeta_0$, low level plans $\bfr$, and dynamic agent variables $\bff$, and associated measurements (shaded circles).}
    \label{fig:single-level-hierarchy}
    \vspace{-5mm}
\end{wrapfigure}Similarly, we represent the low level motion plan with a random trajectory variable $\bfr$ that is governed by the joint distribution $\pigp$ over the platform and environmental agents $\bff=\{\bff^1,\ldots,\bff^{n_t} \}$,  where $\bfz_{1:t} = [\bfz^R_{1:t}, \bfz^f_{1:t}]$ is platform state data (such as localization information) and agent state data respectively (as in~\cite{trautmanicra2013}).  Because the high and low level plans are dependent, we seek models for $\na$, the joint distribution over the high level plan, low level plan, and environmental agents.

In taking a probabilistic approach, our first challenge lies in modeling the distribution in a faithful yet tractable way.  In Figure~\ref{fig:single-level-hierarchy} we present the graphical model of our approach, inspired by the approach taken in~\cite{trautman-smc-2015}; similarly, undirected graphical models of this sort (Markov random fields) have enjoyed great success in the image and natural processing literature.  According to the graphical model in Figure~\ref{fig:single-level-hierarchy}, the distribution factors as
\begin{align*}
\na &= 
\psi_{}(\boeta_0,\bfr)\pzero \pigpshort  \\
&=\psi_{}(\boeta_0,\bfr)\pzero\psi(\bfr,\bff)\probot p(\bff \mid \bfz^f_{1:t})
\end{align*}
where $\psi_{}(\boeta_0,\bfr)\pzero$ encodes the ``agreeability'' of the low level plan and the high level plan and $\pigpshort$ encodes the most likely path through the dynamic, responsive environment (e.g., a crowd of humans).   As was argued in~\cite{trautmaniros}, our model immediately suggests a natural way to perform navigation: at time $t$, find the \emph{maximum a-posteriori} (MAP) assignment for the joint distribution $(\boeta_0, \bfr, \bff)^* = \argmax_{\boeta,\bfr,\bff}\na$, and then take $\bff^{R^*}_{t+1}$ as the next action in the path.  Since $(\boeta_0, \bfr, \bff)^*$ captures the most probable joint value of the high level and low level plan, $\bff^{R^*}_{t+1}$ captures the next actuator command that is most in agreement with the high level plan and most likely to succeed in navigating through the dynamic, responsive environment.

In the following case studies, we present scenarios that move from the most limited form of high level guidance (designating a goal in a map) to a scenario where a human operator interacts arbitrarily (sets as many waypoints as desired/grabs and releases the joystick whenever he pleases).
\subsection{Single global operator instruction}
\begin{wrapfigure}{r}{0.4\textwidth}
\centering \vspace{-1em}
    \includegraphics[width=0.4\textwidth]{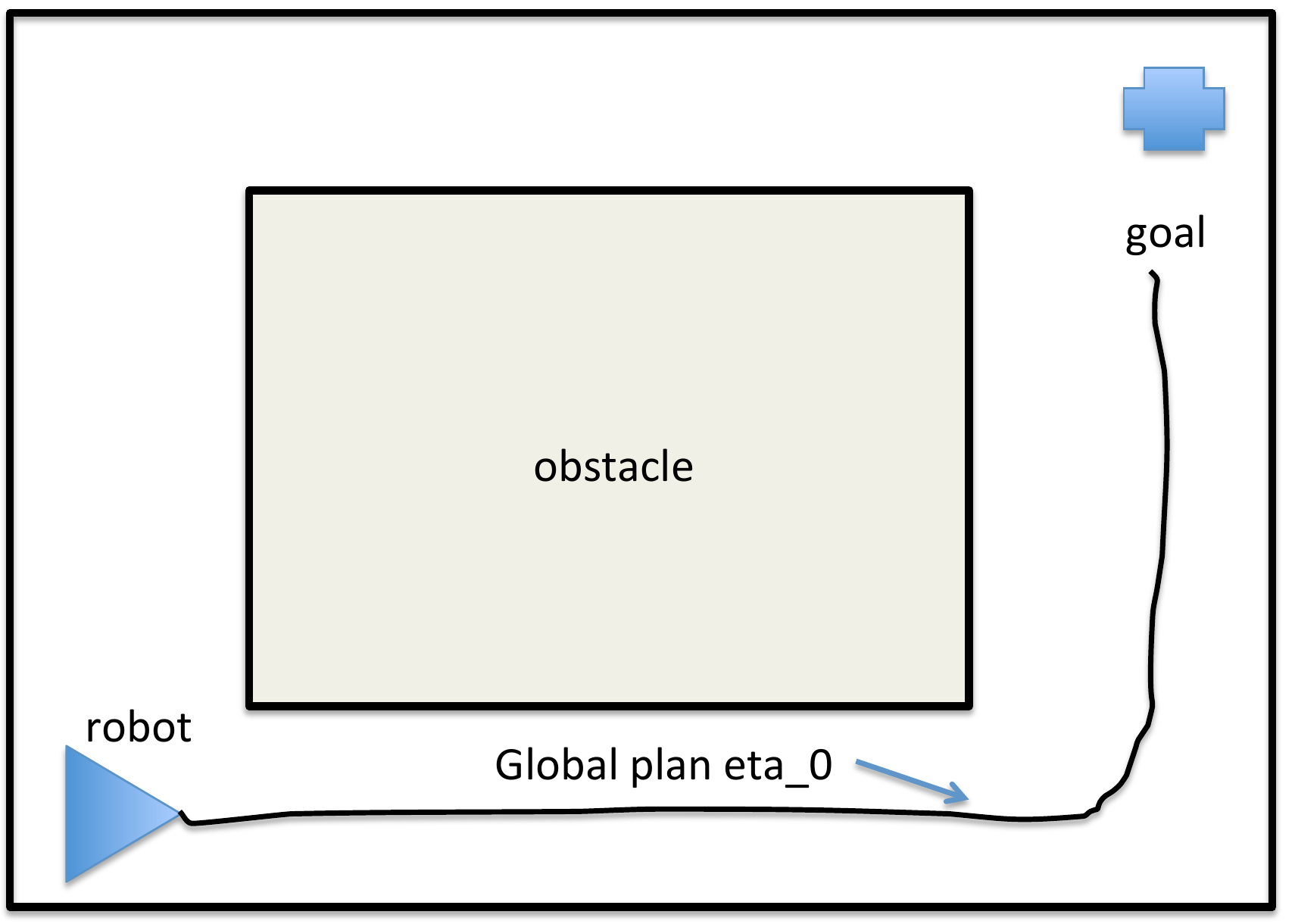}
    \vspace{-1em}
    \caption{Single goal $G$ in a map $m$.}
    \label{fig:gotoG}
    \vspace{-5mm}
\end{wrapfigure}

In Figure~\ref{fig:gotoG} we illustrate the most basic integration of high to low level path planning. 
\begin{itemize} 
\item The environment is static with a single known obstacle.  Thus, we have $\bff = \emptyset$ (i.e., there is no crowd). 
\item The operator designates the goal $G$ in the predefined map $m$. 
\item We can thus utilize a standard global planner; call it A*.  We let $\boeta_0^1 = A^*(m, G)$.  Thus, $p(\boeta_0 \mid G, m) = \delta(\boeta_0 - \boeta_0^1)$.
\item Our local planner is trivial, since there are no local disturbances.
\item We compute ``actuator inputs'' by finding 
\begin{align*}
\bff^{R*} \in (\boeta_0, \bfr, \bff)^* &= \argmax_{\boeta,\bfr,\bff} \psi(\boeta, \bfr)p(\boeta \mid G,m) \pigpshort \\
&=\argmax_{\bfr,\bff} \psi(\boeta_0^1, \bfr)\probot
\end{align*}
where $\probot$ is the kinematic model of the robot, and $\psi(\boeta_0, \bfr)$ is the interaction function between the robot and the global plan.  The operator-robot interaction function $\psi(\boeta_0, \bfr)$ could take a number of forms; we choose $\psi(\boeta_0, \bfr) = \exp(-\frac{1}{2h}[\boeta_0-\bfr][\boeta_0-\bfr]^\top)$.  Thus, $\bff^{R*}$ converges to $\boeta_0^1$ in the situation of Figure~\ref{fig:gotoG}.
\end{itemize}

\paragraph{Relationship to ROS nav stack} Recovering the ROS navigation stack with this approach is trivial: at each time step $t$, sample local paths $\bfr_i \sim \probot$, and weight each sample according to $\psi(\boeta_0^1, \bfr_i)p(\bfr_i \mid \bfz^R_{1:t})$---the first factor encodes global compatibility, while the second factor encodes kinematic feasibility.  Choose the sample $\bfr_i$ with the highest weight as the inputs to the actuators.

The probabilistic formulation allows us to approach the DWA ROS navigation stack in a more general manner: in the ROS navigation stack, sampling from $\probot$ and then weighting amounts to straightforward importance sampling.  However, the distribution $\psi(\boeta_0, \bfr)\probot$ can be approximately inferred using a host of methods: markov chain monte carlo, Laplace approximation, hybrid monte carlo, etc---any approximate inference technique is at our disposal.

In contrast, the ROS navigation stack (\url{http://wiki.ros.org/base_local_planner}), does not pose the high level to low level path planning problem as a probability distribution, so it is not immediately clear how to employ approximate inference techniques to find more accurate solutions in a more efficient manner.

\subsection{Single global operator instruction, multiple static and dynamic obstacles}
\begin{figure*}[th]
\centering
 \subfigure[]{
    \includegraphics[scale=0.33]{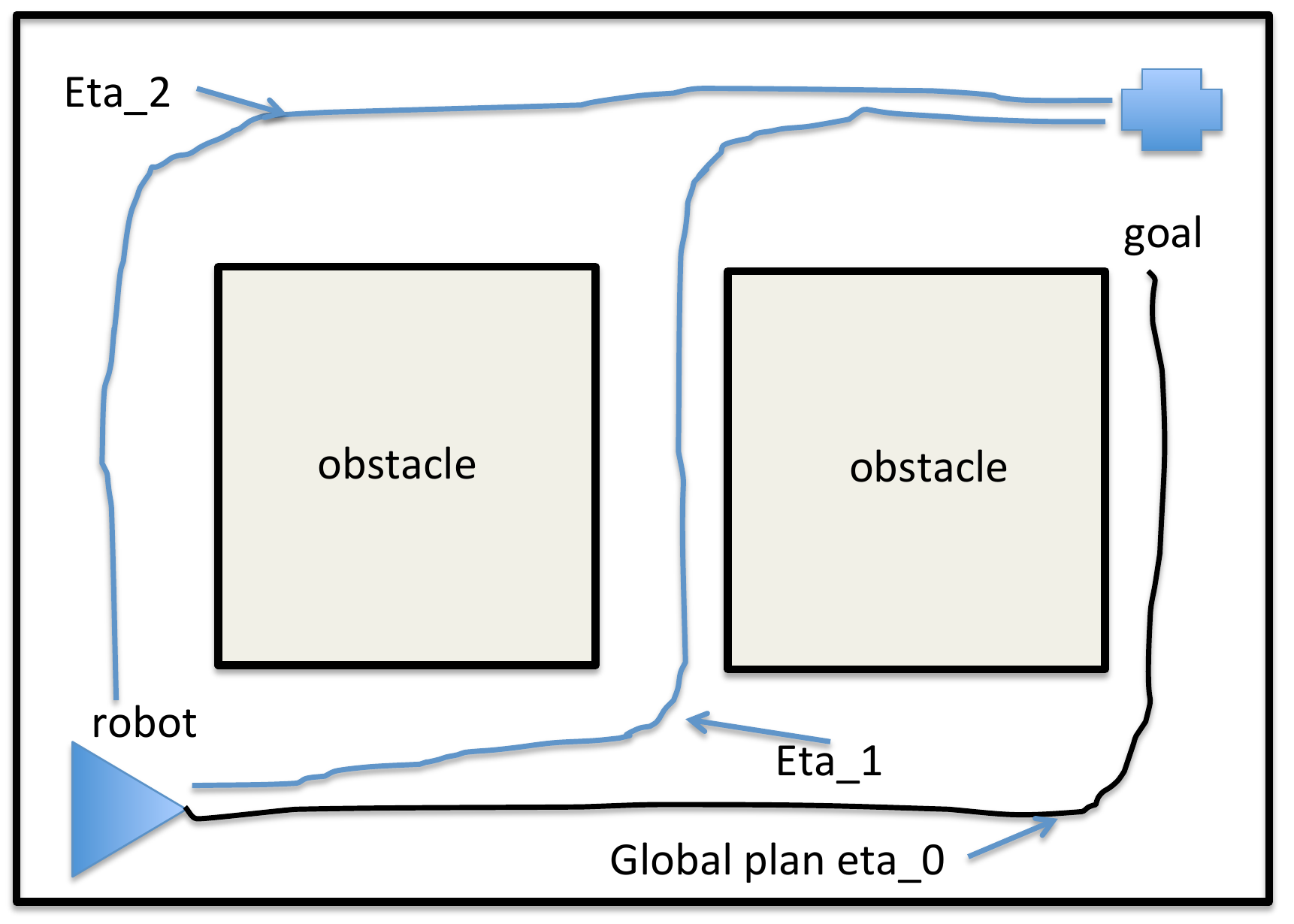}
    \label{fig:gotoG_multiple}
  }
\subfigure[]{
\includegraphics[scale=0.33]{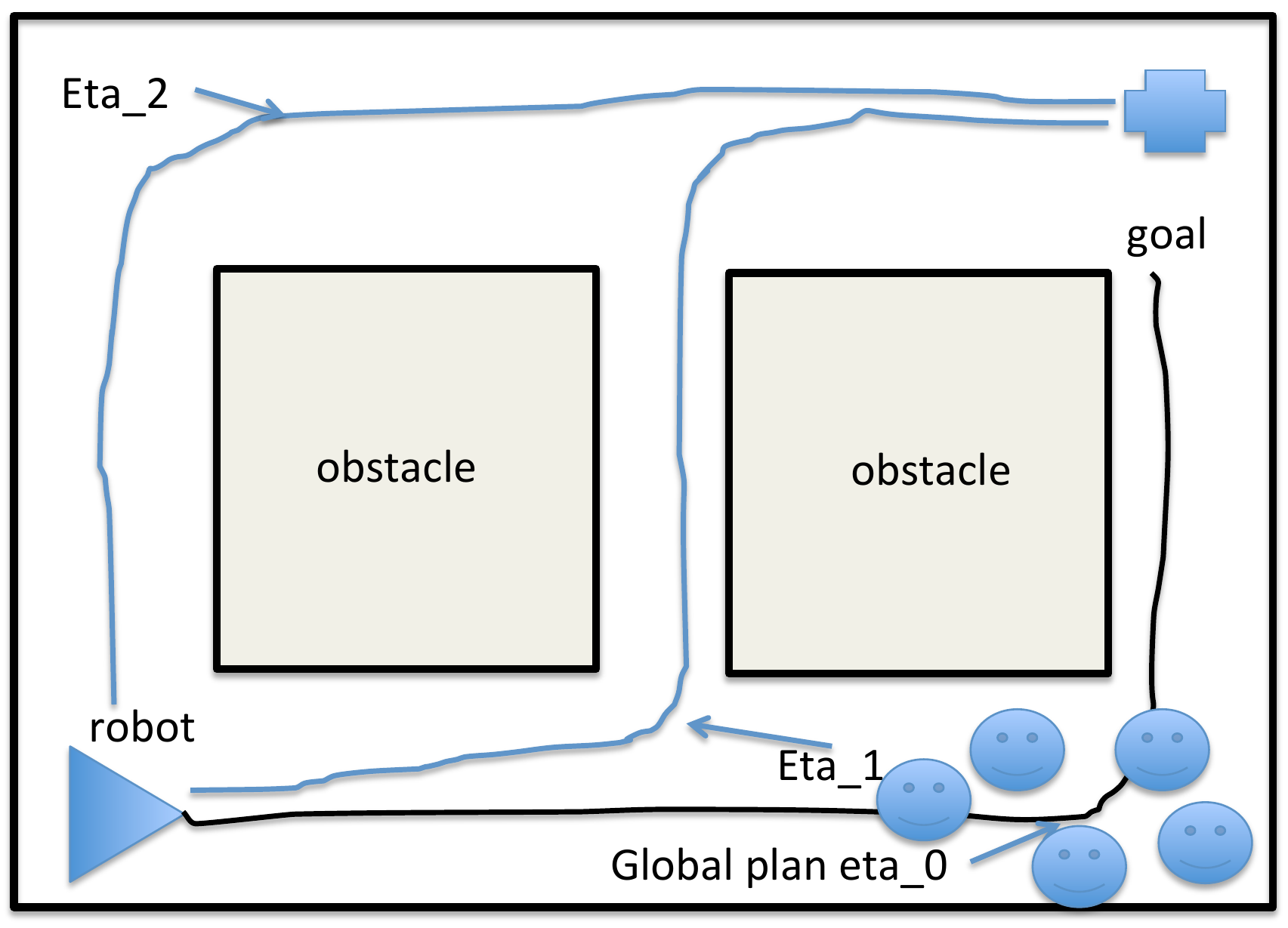}
\label{fig:gotoG_multiple_dynamic}
}
\caption{\textbf{(a)} Global map with two obstacles, and 3 global plans with nontrivial weight \textbf{(b)} Same global map, but 5 smiley faces (5 people) provide a local disturbance in the lower right half of the map. }
\end{figure*}

In Figure~\ref{fig:gotoG_multiple} we introduce the notion of multiple global plans, each of which have nontrivial value.  In particular, global plans $\{\boeta_0^i \}_{i=1}^3$ have values in the static map of $w_0 > w_1 > w_2$.  The global plan distribution thus takes the form
\begin{align*}
p(\boeta_0 \mid G, m) = \sum_{i=1}^3 w_i \delta(\boeta_0-\boeta_0^i).
\end{align*}
In Figure~\ref{fig:gotoG_multiple_dynamic}, we introduce a local crowd disturbance in the bottom right of the map.  We assume that the crowd enters into the robot's field of view near the center corridor;  
thus, the robot has to make a planning decision according to
\begin{align*}
\bff^{R*} \in (\boeta_0, \bfr, \bff)^* &= \argmax_{\boeta_0,\bfr,\bff} \psi(\boeta_0, \bfr)p(\boeta_0 \mid G) \pigpshort \\
&=\argmax_{\boeta_0,\bfr,\bff} \sum_{i=1}^3 w_i \psi(\boeta_0^i,\bfr)\pigpshort.
\end{align*}
When the crowd is not in the robot's field of view, $\bff = \emptyset$, and the low level planner stays close to the optimal global plan $\boeta_0^1$.  However, when $\bff = (\bff^1, \ldots, \bff^5)$, it is no longer obvious which global plan to follow.  With our probabilistic approach, which global path to follow is determined by balancing the capabilities of the low level planner in the crowd (effectively, how much probability is in $\pigpshort$ near the global plan $\boeta_0^i$) against how much more efficient $\boeta_0^1$ is than $\boeta_0^2$ (or, how $w_1$ compares to $w_2$).

Heuristically, one can think of the distribution $ \sum_{i=1}^3 w_i \psi(\boeta_0^i,\bfr)\pigpshort$ as having three modes, each (roughly) centered around the global plans.  The relative probability mass in each mode ends up determining the MAP value of the distribution.  Thus, the global plan's fitness---represented by $w_i$---is balanced against the challenge of the local situation, which is represented by $\pigpshort$.
\subsection{Global goal specified, operator intervenes randomly }
\begin{wrapfigure}{r}{0.51\textwidth}
\centering \vspace{-1em}
    \includegraphics[width=0.5\textwidth]{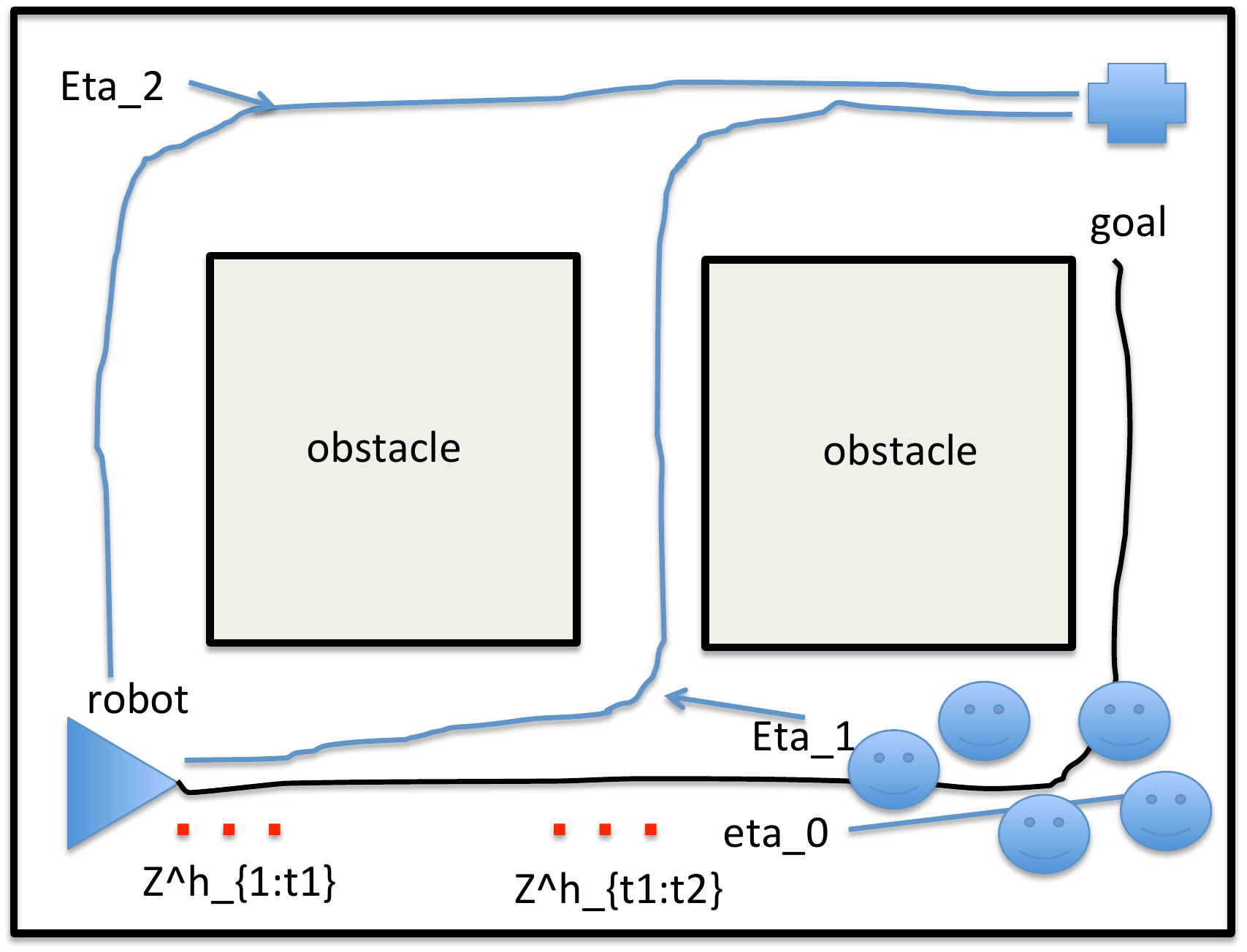}
    \vspace{-1em}
    \caption{Single goal $G$ in a map $m$ with local disturbances.  The operator intervenes at $\bfz^h_{1:t_1}$ and $\bfz^h_{t_2:t_3}$ (red dots). }
    \label{fig:intervene}
\end{wrapfigure}

Now, suppose that the operator has provided a global goal $G$ (and thus high level plans are generated), but intervenes via a joystick at random times according to  $\bfz^h_{1:t_1}$ and $\bfz^h_{t_2:t_3}$, as in Figure~\ref{fig:intervene} (the difference between this scenario and the scenario in Figure~\ref{fig:gotoG_multiple_dynamic} is the presence of the joystick data). 
 Now, not only do we have to balance global considerations (the weights of the global plans) against local disturbances, but also the online desires of the operator.  In particular, the robot will move through the environment in the same manner as in Figure~\ref{fig:gotoG_multiple_dynamic}, until the operator intervenes with the joystick at $\bfz^h_{1:t_1}$.  At this point, the global plan distribution will become $p(\boeta_0 \mid G, \bfz^h_{1:t_1})$, and thus influence local decision making by ``pulling'' $\pigpshort$ towards the more peaked regions of $p(\boeta_0 \mid G, \bfz^h_{1:t_1})$---we are able to simultaneously represent high level operator desires with online refinements.  Our full joint distribution now becomes $p(\boeta_0, \bfr, \bff \mid G, \bfz_{1:t},\bfz^h_{1:t_1})$.

\paragraph{High/low level plan arbitration and assistive technologies}Importantly, in the absence of a global goal $G$, the formulation reduces to $p(\boeta_0, \bfr, \bff \mid \bfz_{1:t},\bfz^h_{1:t_1})$.  This is the case of fully assistive shared control, where the absence of a global map or corrupted localization data renders the global goal $G$ meaningless.  In this case, the global plan $\boeta_0$ is revealed incrementally via local user input data $\bfz^h_{1:t_1}$.  This capability becomes important when, for instance, the robot enters a crowd, and standard localization techniques start to fail---at this point the robot must ``share awareness'' with the operator by inferring global destinations from local operator input data.

\paragraph{Probabilistic factorizations guided by formal methods}
\begin{wrapfigure}{r}{0.5\textwidth}
\centering \vspace{-1em}
    \includegraphics[width=0.5\textwidth]{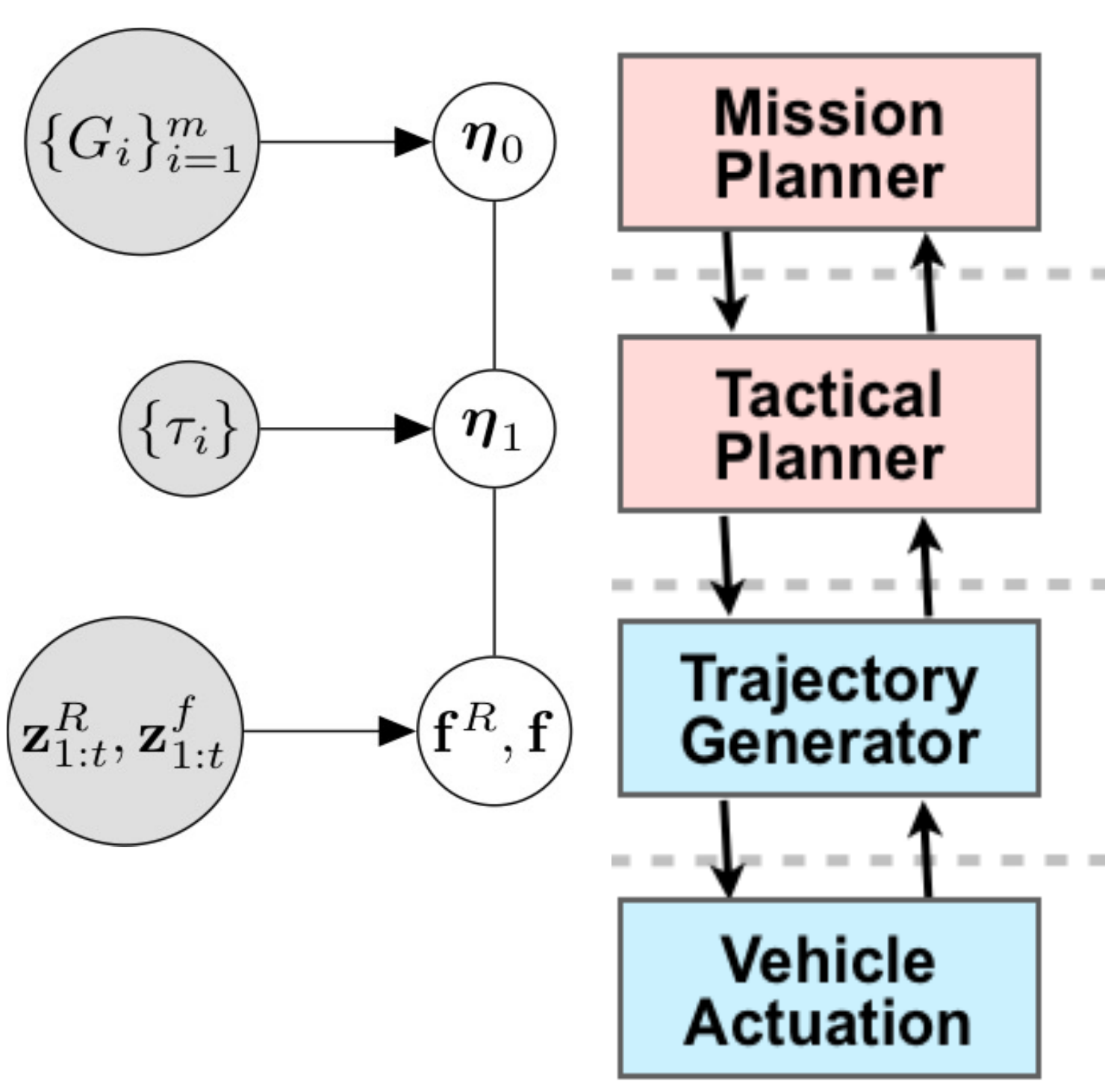}
    \vspace{-1em}
    \caption{Comparison of graphical model decomposition and a formal hierarchical decision stack. Right half of image courtesy of Ufuk Topcu.}
    \label{fig:pgm-to-hier}
\end{wrapfigure}
While the success of previous experiments and the utility of the Markov random field factorization  lend credence to our model above, we point out that results from formal methods (and thus provably correct constructions) can guide how we model our joint distribution (courtesy of Ufuk Topcu).  To see how, we refer to Figure~\ref{fig:pgm-to-hier}, where we have plotted the state of the art formal methods decision stack next to its corresponding graphical model decomposition.  Note that the results from formal methods suggests that a ``tactical variable'', which we call $\boeta_1$, is used to mediate information between the high level $\boeta_0$ and the low level $\bfr$ (we assume that some form of tactical data, $\{\tau_i \}_{i=1}^k$ informs the governing distribution $p(\boeta_1 \mid \{\tau_i \}_{i=1}^k)$). This graphical model in turn represents the factorization

\begin{align*}
p(\boeta_0, \boeta_1,\bfr,\bff\mid \bfz_{1:t}, \{\tau_i \}_{i=1}^k,\{G_i \}_{i=1}^m) &=  \psi(\boeta_0, \boeta_1) p(\boeta_0\mid \{G_i \}_{i=1}^m)\times  \\
&\psi(\boeta_1,\bfr)p(\boeta_1 \mid \{\tau_i \}_{i=1}^k) \pigpshort.
\end{align*}
One of our ongoing research objectives is to fully understand how formal methods can guide our probabilistic decompositions---while the probabilistic approach is well suited to capture dependencies between variables and flexible enough to capture the vagaries of human behavior (or online manipulation of robot objectives), balancing tractability and fidelity in the factorization of the joint distribution is more of an anecdotal art than a science.  Results from formal analysis, however, can provide guidance on our decomposition and potentially insight into the form of our\begin{wrapfigure}{r}{0.51\textwidth}
\centering \vspace{-1em}
    \includegraphics[width=0.5\textwidth]{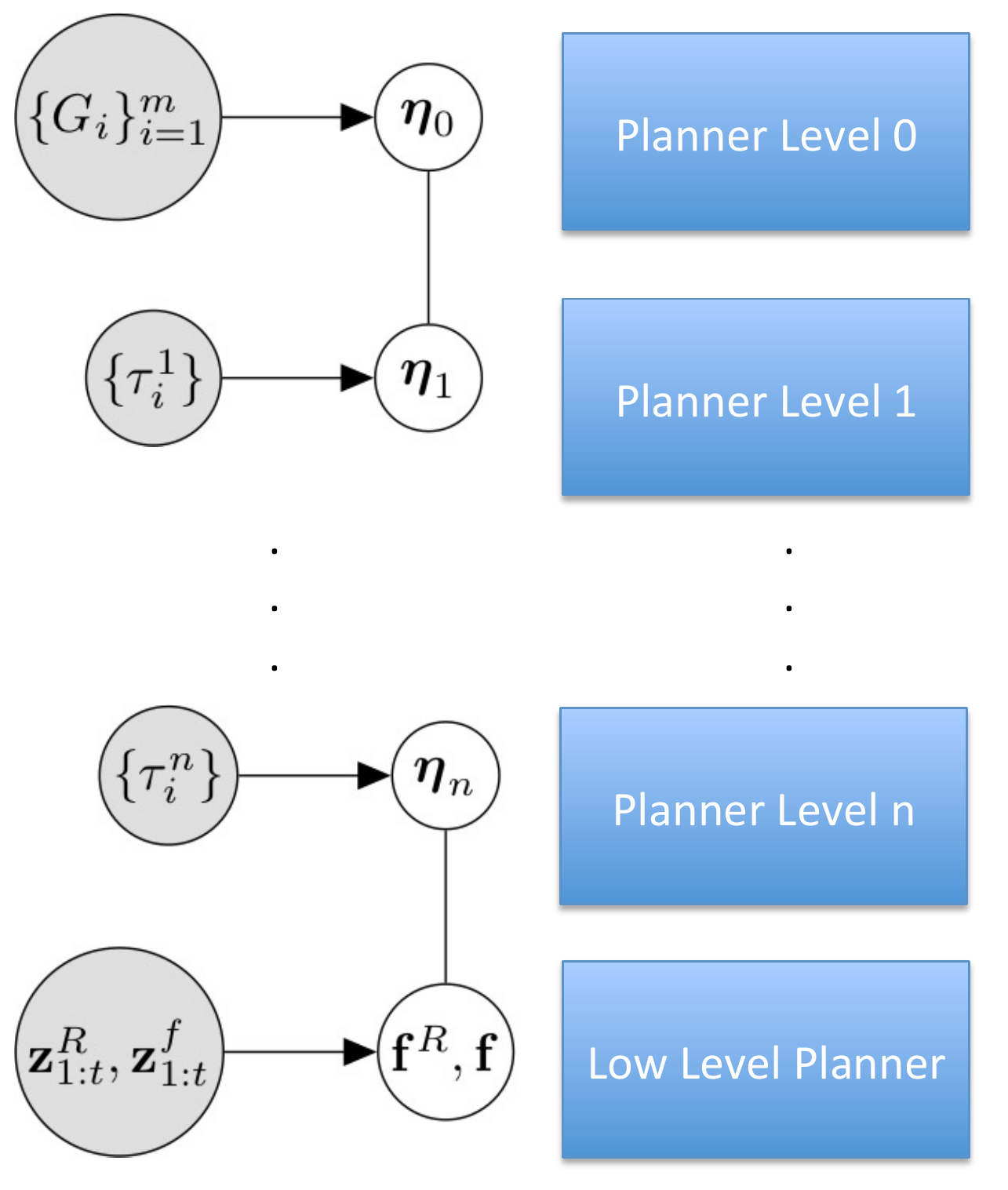}
    \vspace{-1em}
    \caption{Comparison of graphical model decomposition and an arbitrarily sized formal hierarchical decision stack.  Each planning level determined by formal analysis determines a corresponding planning variable in the probabilistic formulation.  }
    \label{fig:arbitrary-stacks}
\end{wrapfigure} ``cooperation functions'' $\psi(\boeta_0, \boeta_1)$ and $\psi(\boeta_1,\bfr)$  that link the mission, tactical, and trajectory levels.  Furthermore, it is not immediately clear how to relate data coming in at various levels to planning level variables (e.g., high level symbolic data $\{G_i \}$ clearly relates to $\boeta_0$; however, the introduction of other planning levels necessitates understanding of how lower level data---such as joystick commands in the form of $\bfz^h_{1:t}$---measures lower level planning variables).

\paragraph{Arbitrary decision stack factorizations}
The hierarchical decision stack illustrated in Figure~\ref{fig:pgm-to-hier} was tied to a specific application, and so is not in general the correct hierarchical decomposition.  However, the approach of finding provably correct hierarchical decompositions for arbitrary scenarios, and then reading off the corresponding graphical model (and thus probabilistic decomposition) is fully general; we depict this approach in Figure~\ref{fig:arbitrary-stacks}.  In combination with human-learning and symbolic planner approaches (which guide how we model and adapt distributions, such as $p(\boeta_0 \mid \{G_i \}_{i=1}^m)$ and $p(\boeta_k \mid \{ \tau_i^k \}_{i=1}^m)$, at specific levels of the planning stack), our approach has the potential to be both flexible enough to accommodate a wide variety of online manipulation of global robot objectives while maintaining the rigor of formal analysis.

\bibliographystyle{apalike}
{\footnotesize
\bibliography{standard_bibliography}

\begin{thebibliography}{}

\bibitem[Trautman, 2015]{trautman-smc-2015}
Trautman, P. (2015).
\newblock Assistive planning in complex, dynamic environments.
\newblock In {\em IEEE Systems, Man, and Cybernetics
  (http://arxiv.org/abs/1506.06784)}.

\bibitem[Trautman and Krause, 2010]{trautmaniros}
Trautman, P. and Krause, A. (2010).
\newblock Unfreezing the robot: Navigation in dense interacting crowds.
\newblock In {\em IROS}.

\bibitem[Trautman et~al., 2013]{trautmanicra2013}
Trautman, P., Ma, J., Krause, A., and Murray, R.~M. (2013).
\newblock Robot navigation in dense crowds: the case for cooperation.
\newblock In {\em ICRA}.

\end{thebibliography}
}

\end{document}